\documentclass[10pt,twocolumn,letterpaper]{article}

\usepackage{cvpr}
\usepackage{times}
\usepackage{epsfig}
\usepackage{graphicx}
\usepackage{amsmath}
\usepackage{amssymb}

\usepackage{helvet}  
\usepackage{courier}  
\usepackage{url}  
\usepackage{color}
\usepackage{amsfonts}
\usepackage{bbm}
\usepackage{float}
\usepackage[ruled,lined]{algorithm2e}
\usepackage{epsfig}
\usepackage{booktabs}

\usepackage[pagebackref=true,breaklinks=true,letterpaper=true,colorlinks,bookmarks=false]{hyperref}

\cvprfinalcopy 

\def\cvprPaperID{343} 

\ifcvprfinal\fi
\begin{document}

\title{Generative Partition Networks for Multi-Person Pose Estimation}

\author{
  Xuecheng Nie$^1$, Jiashi Feng$^1$, Junliang Xing$^2$, Shuicheng Yan$^{1,3}$\\
  $^1$ECE Department, National University of Singapore, Singapore\\
  $^2$Institute of Automation, Chinese Academy of Sciences, Beijing, China\\
  $^3$Qihoo 360 AI Institute, Beijing, China \\
  \footnotesize{\texttt{niexuecheng@u.nus.edu, elefjia@nus.edu.sg, jlxing@nlpr.ia.ac.cn, yanshuicheng@360.cn}} \\
}

\maketitle

\begin{abstract}
   This paper proposes a new Generative Partition Network (GPN) to address the challenging multi-person pose estimation problem. Different from existing  models that are either completely
   top-down or bottom-up, the proposed GPN introduces a novel strategy\textemdash it generates partitions for multiple persons from their global joint candidates and infers instance-specific
   joint configurations \emph{simultaneously}.
   The GPN is favorably featured by low complexity and high accuracy of joint detection and re-organization.
   In particular, GPN designs a generative model that performs one feed-forward pass to efficiently generate robust person detections with joint partitions, relying on dense regressions from
   global joint candidates in an embedding space parameterized by centroids of persons.
   In addition, GPN formulates the inference procedure for joint configurations of human poses as a graph partition problem, and conducts local optimization for each person
   detection with reliable global affinity cues, leading to complexity reduction and performance improvement. GPN is implemented with the Hourglass architecture as the backbone network to
   simultaneously learn joint detector and dense regressor.
   Extensive experiments on benchmarks MPII Human Pose Multi-Person, extended PASCAL-Person-Part, and WAF, show the efficiency of GPN with new state-of-the-art performance.
\end{abstract}

\section{Introduction}

\begin{figure}[t!]
\begin{center}
\setlength{\tabcolsep}{1.5pt}
\begin{tabular}{ccc}
\includegraphics[scale=0.56]{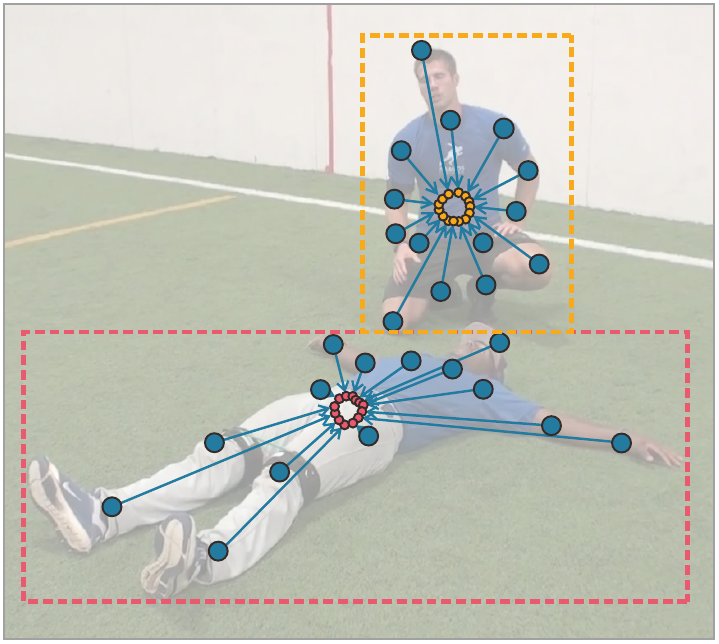} & \includegraphics[scale=0.56]{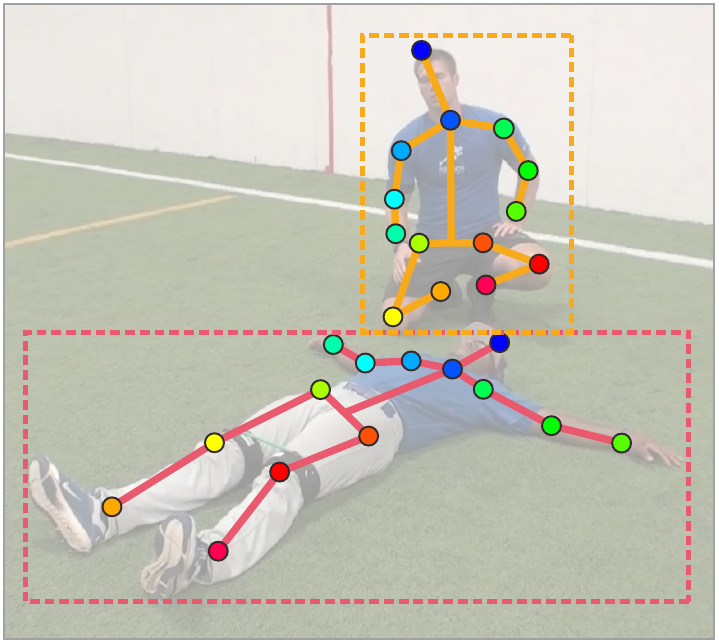} \\
{\fontsize{9pt}{9pt} \selectfont (a) Generative Partition} & {\fontsize{9pt}{9pt} \selectfont (b) Local Inference} \\
\end{tabular}
\vspace{0.5mm}
\caption{Generative Partition Networks for multi-person pose estimation. (a) Generative partition. GPN models person detection and joint partition as a generative process inferred from joint
candidates. (b) Local inference. GPN performs local inference for joint configurations conditioned on the generated person detections with joint partitions.}
\label{fig:fig1}
\end{center}
\end{figure}

Multi-person pose estimation aims to localize body joints of multiple persons captured in a 2D monocular image~\cite{eichner2010we,hpe:deepcut_cvpr16}. Despite extensive prior research, this problem  remains very challenging due to the highly complex joint configuration,  partial or even complete joint occlusion, significant overlap between neighboring persons, unknown number of persons and more critically the difficulties in allocating   joints to multiple persons.
These challenges feature the unique property of multi-person pose estimation compared with the simpler single-person setting~\cite{hpe:hourglass_arxiv15,hpe:conv_pose_machine_arxiv16}. To tackle these challenges, existing multi-person pose estimation approaches usually perform joint detection and  partition \emph{separately}, mainly following two different strategies. The \emph{top-down}
strategy~\cite{eichner2010we,fang16rmpe,iqbal2016multi,papandreou2017towards,pishchulin2012articulated} first detects persons and then performs pose estimation for each single person individually.
The \emph{bottom-up} strategy~\cite{cao2017realtime,insafutdinov2016articulated,hpe:deepercut_eccv16,ladicky2013human,levinkov2017joint,hpe:deepcut_cvpr16}, in contrast, generates all joint candidates
at first, and then tries to partition them to corresponding person instances.

The top-down approaches  directly leverage existing   person detection models~\cite{liu2016ssd,ren2015faster} and single-person pose estimation  methods~\cite{hpe:hourglass_arxiv15,hpe:conv_pose_machine_arxiv16}. Thus they effectively avoid complex  joint partitions.
However, their performance is critically limited by the quality of person detections. If the employed person detector fails to detect a person instance accurately (due to occlusion, overlapping or other distracting factors), the introduced errors cannot be remedied  and would severely harm performance of the following pose estimation. Moreover, they suffer from high  joint detection complexity, which linearly increases with the number of persons in the image, because they need  to run the single-person joint detector for each person detection sequentially.

In contrast, the bottom-up approaches  detect all joint candidates at first by globally applying a joint  detector for only once and then partition them to corresponding persons according to joint affinities. Hence, they enjoy lower joint detection complexity than the top-down ones and  robustness to errors from early commitment. However, they suffer from very high complexity of partitioning joints to corresponding persons, which usually involves solving NP-hard graph partition problems~\cite{hpe:deepercut_eccv16,hpe:deepcut_cvpr16} on densely connected graphs covering the whole image.

In this paper, we propose a novel solution, termed the \emph{Generative Partition Network} (GPN),  to overcome essential limitations of the
above two types of approaches and meanwhile inherit their strengths within a unified model for efficiently and effectively estimating poses of multiple persons in a given image.
As shown in Figure~\ref{fig:fig1}, GPN solves multi-person pose estimation problem by simultaneously
1) modeling person detection for joint partition as a generative process inferred from all joint candidates and
2) performing local inference for obtaining joint categorizations and associations conditioned on the generated person detections.

In particular, GPN introduces a dense regression module to generate person detections with partitioned joints via votes from joint candidates in a carefully designed embedding space, which is efficiently parameterized by person centroids. This generative partition model produces joint candidates and partitions by running a joint detector for only one feed-forward pass, offering much higher efficiency than top-down approaches. In addition, the produced person detections from GPN are robust to various distracting factors, \eg, occlusion, overlapping, deformation, and large pose variation, benefiting the following pose estimation. GPN also introduces a local greedy inference algorithm by assuming independence among person detections for producing optimal multi-person joint configurations. This local optimization strategy reduces the search space of the graph partition problem for finding optimal poses, avoiding the high joint partition complexity challenging the bottom-up strategy. Moreover, the local greedy inference algorithm exploits reliable global affinity cues from the embedding space for inferring joint configurations within robust person detections, leading to performance improvement.


We implement GPN based on the Hourglass network~\cite{hpe:hourglass_arxiv15} for learning joint detector and dense regressor, simultaneously. Extensive experiments on MPII Human Pose Multi-Person~\cite{andriluka14cvpr}, extended PASCAL-Person-Part~\cite{xia2017joint} and WAF~\cite{eichner2010we} benchmarks evidently show the efficiency and effectiveness of the proposed GPN. Moreover, GPN achieves new state-of-the-art on all these benchmarks.

We make following contributions.
1) We propose a new one-pass solution to multi-person pose estimation, totally different from previous top-down and bottom-up ones.
2) We propose a novel dense regression module to efficiently and robustly partition body joints into multiple persons, which is the key to speeding up multi-person pose estimation.
3) In addition to high efficiency,  GPN is also superior in terms of robustness and accuracy on multiple benchmarks.

\section{Related Work}

\vspace{-2mm}
\paragraph{Top-Down Multi-Person Pose Estimation}
Existing approaches following top-down strategy sequentially perform person detection and
single-person pose estimation. In~\cite{gkioxari2014using}, Gkioxari \emph{et al.} proposed to adopt the Generalized Hough Transform framework to first generate person proposals and then classify joint candidates based on the poselets. Sun \emph{et al.}~\cite{sun2011articulated} presented a hierarchical part-based model for jointly person detection and pose estimation. Recently, deep learning
techniques have been exploited to improve both person detection and single-person pose estimation. In~\cite{iqbal2016multi}, Iqbal and Gall adopted Faster-RCNN~\cite{ren2015faster} based person detector
and convolutional pose machine~\cite{hpe:conv_pose_machine_arxiv16} based joint detector for this task. Later, Fang \emph{et al.}~\cite{fang16rmpe} utilized spatial transformer network~\cite{jaderberg2015spatial}
and Hourglass network~\cite{hpe:hourglass_arxiv15} to further improve the quality of joint detections and partitions. Despite remarkable success, they suffer from limitations
from early commitment and  high joint detection complexity. Differently, the proposed GPN adopts a one-pass generative process for efficiently producing person detections with partitioned joint candidates, offering robustness to  early commitment as well as low joint detection complexity.

\vspace{-4mm}
\paragraph{Bottom-Up Multi-Person Pose Estimation} The bottom-up strategy provides robustness to early commitment and low joint detection complexity. Previous
bottom-up approaches~\cite{cao2017realtime,hpe:deepercut_eccv16,newell2016associative,hpe:deepcut_cvpr16} mainly focus on improving either the joint detector or  joint affinity cues, benefiting the following
joint partition and configuration inference. For joint detector, fully convolutional neural networks, \emph{e.g.},
Residual networks~\cite{he2016deep} and Hourglass networks~\cite{hpe:hourglass_arxiv15}, have been widely exploited. As for joint affinity cues, Insafutdinov \emph{et al.}~\cite{hpe:deepercut_eccv16}
explored geometric and appearance constraints among joint candidates. Cao \emph{et al.}~\cite{cao2017realtime} proposed part affinity fields to encode location and orientation of limbs.
Newell and Deng~\cite{newell2016associative} presented the associative embedding for grouping joint candidates. Nevertheless, all these approaches partition joints  based on partitioning the
graph covering the whole image, resulting in high inference complexity. In contrast, GPN performs local inference with
robust global affinity cues which is efficiently  generated by dense regressions from the centroid embedding space, reducing complexity  for joint partitions and improving pose estimation.

\begin{figure*}[t!]
\begin{center}
\includegraphics[scale=0.58]{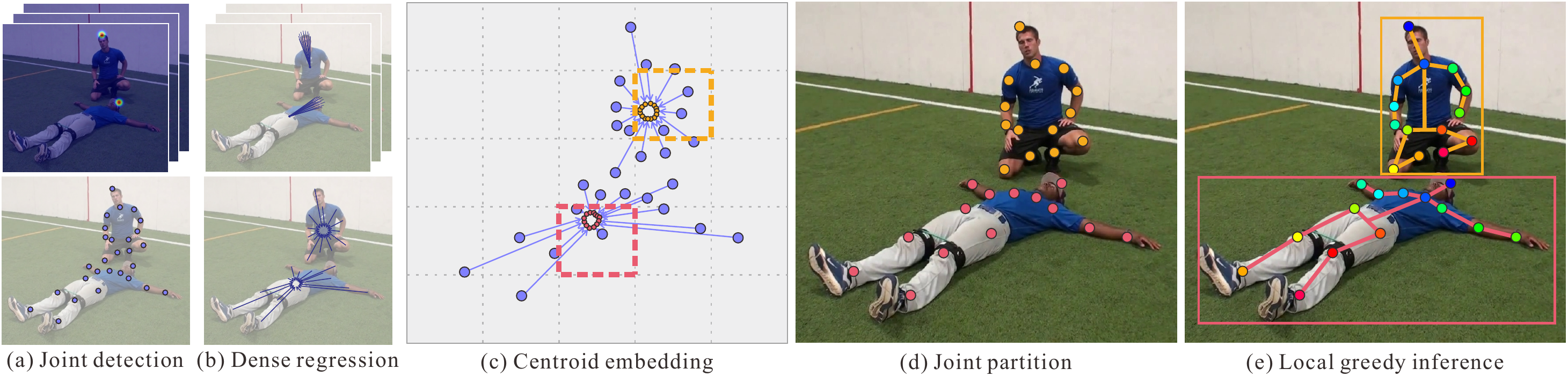}
\caption{Overview of the proposed Generative Partition Network for multi-person pose estimation. Given an image, GPN first uses a CNN to predict (a) joint confidence maps and (b) dense joint-centroid regression maps.
         Then, GPN performs (c) centroid embedding for all joint candidates in the embedding space via dense regression, to produce (d) joint partitions within person detections.
         Finally, GPN conducts (e) local greedy inference to generate joint configurations for each joint partition locally, giving pose estimation results of multiple persons.}
\label{fig:flowchart}
\end{center}
\vspace{-20pt}
\end{figure*}

\section{Approach}

\subsection{Generative Partition Model}

The overall pipeline for the proposed  Generative Partition Network (GPN) model is shown in Figure~\ref{fig:flowchart}. Throughout the paper, we use following notations. Let  $\mathbf{I}$ denote an image containing multiple persons, $\mathbf{p}{=}\{\mathbf{p}_1,\mathbf{p}_2,\ldots, \mathbf{p}_N\}$ denote the spatial coordinates  of $N$ joint candidates from all persons in $\mathbf{I}$ with $\mathbf{p}_v{=}(x_v,y_v)^\top, \forall v{=}1,\ldots,N$, and $\mathbf{u}{=}\{u_1, u_2,\ldots,u_N\}$ denote the labels of corresponding joint candidates, in which $u_v{\in}\{1, 2, \ldots, K\}$ and  $K$  is the number of joint categories. For allocating joints via local inference, we also consider the proximities  between  joints, denoted as $\mathbf{b}{\in}\mathbb{R}^{N \times N} $. Here $\mathbf{b}_{(v,w)}$ encodes the proximity  between  the $v$th joint candidate $(\mathbf{p}_v, u_v)$
and the $w$th joint candidate $(\mathbf{p}_w, u_w)$, and gives the probability for them to be from the same person.

The proposed GPN  with learnable parameters $\Theta$ aims  to  solve the multi-person pose estimation task through learning to infer the conditional distribution $\mathbb{P}(\mathbf{p}, \mathbf{u}, \mathbf{b} | \mathbf{I}, \Theta)$. Namely,  given the image $\mathbf{I}$, GPN infers the joint locations $\mathbf{p}$, labels $\mathbf{u}$ and proximities $\mathbf{b}$ providing the largest likelihood probability. To this end,  GPN adopts a generative model to \emph{simultaneously} produce person detections with joint partitions implicitly and infers joint configuration $\mathbf{p}$ and  $\mathbf{u}$ for each person detection locally. In this way, GPN reduces the difficulty and complexity of multi-person pose estimation significantly. Formally, GPN introduces latent variables  $\mathbf{g}{=}\{\mathbf{g}_1,\mathbf{g}_2,\ldots,\mathbf{g}_M\}$ to encode joint partitions, and each $\mathbf{g}_i$ is a collection of joint candidates belonging  to a specific person detection in which their labels are not considered, and $M$ is the number of joint partitions.
With these latent variables $\mathbf{g}$, $\mathbb{P}(\mathbf{p}, \mathbf{u}, \mathbf{b} | \mathbf{I}, \Theta)$ can be factorized  into
\begin{equation}\label{eq:basic_model}
\begin{aligned}
\mathbb{P}(\mathbf{p}, \mathbf{u}, \mathbf{b} | \mathbf{I}, \Theta) & = \sum_{\mathbf{g}}\mathbb{P}(\mathbf{p}, \mathbf{u}, \mathbf{b}, \mathbf{g} | \mathbf{I}, \Theta) \\
& \hspace{-16mm } = \sum_{\mathbf{g}} \underbrace{\mathbb{P}(\mathbf{p}|\mathbf{I}, \Theta)\mathbb{P}(\mathbf{g}|\mathbf{I}, \Theta, \mathbf{p})}_{\text{partition generation}} \underbrace{\mathbb{P}(\mathbf{u}, \mathbf{b}|\mathbf{I}, \Theta, \mathbf{p}, \mathbf{g})}_{\text{joint configuration}},
\end{aligned}
\end{equation}
where $\mathbb{P}(\mathbf{p}|\mathbf{I}, \Theta)\mathbb{P}(\mathbf{g}|\mathbf{I}, \Theta, \mathbf{p})$ models the generative process of joint partitions within person detections based on joint candidates. Maximizing the above likelihood probability  gives optimal pose estimation for multiple persons in  $\mathbf{I}$.

However, directly maximizing the above  likelihood is computationally intractable.  Instead of maximizing \emph{w.r.t.}\ all possible partitions $\mathbf{g}$, we propose to maximize its lower bound
induced by a single ``optimal'' partition, inspired by the EM algorithm~\cite{dempster1977maximum}. Such approximation  could reduce the complexity significantly without harming the performance.
Concretely,  based on Eqn.~(\ref{eq:basic_model}), we have
\begin{equation*}
\begin{aligned}
\mathbb{P}(\mathbf{p}, \mathbf{u}, \mathbf{b} | \mathbf{I}, \Theta) & \geq \\
& \hspace{-18mm} \mathbb{P}(\mathbf{p}|\mathbf{I}, \Theta)\left\{\max_{\mathbf{g}}\mathbb{P}(\mathbf{g}|\mathbf{I}, \Theta, \mathbf{p})\right\}\mathbb{P}(\mathbf{u}, \mathbf{b}|\mathbf{I}, \Theta, \mathbf{p}, \mathbf{g}).
\end{aligned}
\end{equation*}
Here, we find the optimal solution by maximizing the above induced lower bound  $\mathbb{P}(\mathbf{p}, \mathbf{u}, \mathbf{b}, \mathbf{g} | \mathbf{I}, \Theta)$, instead of maximizing the summation.
The joint partitions $\mathbf{g}$ disentangle independent joints and reduce inference complexity\textemdash  only the joints falling in the same partition have non-zero proximities $ \mathbf{b} $. Then $\mathbb{P}(\mathbf{p}, \mathbf{u}, \mathbf{b}, \mathbf{g}| \mathbf{I}, \Theta)$ is further factorized as
\begin{equation}\label{eq:factor_model}
\begin{aligned}
\mathbb{P}(\mathbf{p}, \mathbf{u}, \mathbf{b}, \mathbf{g} | \mathbf{I}, \Theta) & = \mathbb{P}(\mathbf{p}, \mathbf{g}|\mathbf{I}, \Theta)  \\
& \hspace{-20mm} \times \prod_{\mathbf{g}_i \in \mathbf{g}}  \mathbb{P}(\mathbf{u}_{\mathbf{g}_i}|\mathbf{I}, \Theta, \mathbf{p}, \mathbf{g}_i) \mathbb{P}(\mathbf{b}_{\mathbf{g}_i}|\mathbf{I}, \Theta, \mathbf{p}, \mathbf{g}_i, \mathbf{u}),
\end{aligned}
\end{equation}
where $\mathbf{u}_{\mathbf{g}_i}$ denotes the labels of joints falling in the partition $\mathbf{g}_i$ and $\mathbf{b}_{\mathbf{g}_i}$ denotes their  proximities. In the above probabilities, we define $\mathbb{P}(\mathbf{p}, \mathbf{u}, \mathbf{b}, \mathbf{g} | \mathbf{I}, \Theta)$ as a Gibbs distribution:
\begin{equation}\label{eq:gibbs_distribution}
\mathbb{P}(\mathbf{p}, \mathbf{u}, \mathbf{b}, \mathbf{g} | \mathbf{I}, \Theta) \propto \exp\{-E(\mathbf{p}, \mathbf{u}, \mathbf{b}, \mathbf{g})\},
\end{equation}
where $E(\mathbf{p}, \mathbf{u}, \mathbf{b}, \mathbf{g})$ is the energy function for the joint distribution $\mathbb{P}(\mathbf{p}, \mathbf{u}, \mathbf{b}, \mathbf{g} | \mathbf{I}, \Theta)$. Its explicit form is derived from Eqn.~(\ref{eq:factor_model}) accordingly:
\begin{equation}\label{eq:energy_func}
\begin{aligned}
& E(\mathbf{p}, \mathbf{u}, \mathbf{b}, \mathbf{g})  =  -  {\varphi(\mathbf{p}, \mathbf{g})} \\
& \hspace{4mm} - \sum_{\mathbf{g}_i \in \mathbf{g}}\Big( { \sum_{\mathbf{p}_v \in \mathbf{g}_i} \hspace{-2mm} \psi(\mathbf{p}_v, u_v)} + \hspace{-2mm} {\sum_{\mathbf{p}_v,\mathbf{p}_w \in \mathbf{g}_i} \hspace{-4mm} \phi(\mathbf{p}_v, u_v, \mathbf{p}_w, u_w)}\Big).
\end{aligned}
\end{equation}
Here, $\varphi(\mathbf{p},\mathbf{g})$ scores the quality of joint partitions  $\mathbf{g}$ generated from joint candidates $\mathbf{p}$ for the input image $ \mathbf{I} $, $\psi(\mathbf{p}_v, u_v)$
scores how the position $\mathbf{p}_v$ is compatible with label $u_v$, and $\phi(\mathbf{p}_v, u_v, \mathbf{p}_w, u_w)$ represents how likely the positions $\mathbf{p}_v$ with label $u_v$ and $\mathbf{p}_w$ with label $u_w$ belong to the same person, \emph{i.e.}, characterizing the proximity $\mathbf{b}_{(v,w)}$.
In the following subsections, we will give  details for detecting joint candidates $ \mathbf{p} $, generating optimal joint partitions $ \mathbf{g} $, inferring joint configurations $ \mathbf{u} $ and $ \mathbf{b} $ along with the proposed algorithm to optimize the energy function.

\begin{figure}[t!]
\begin{center}
\includegraphics[scale=0.5]{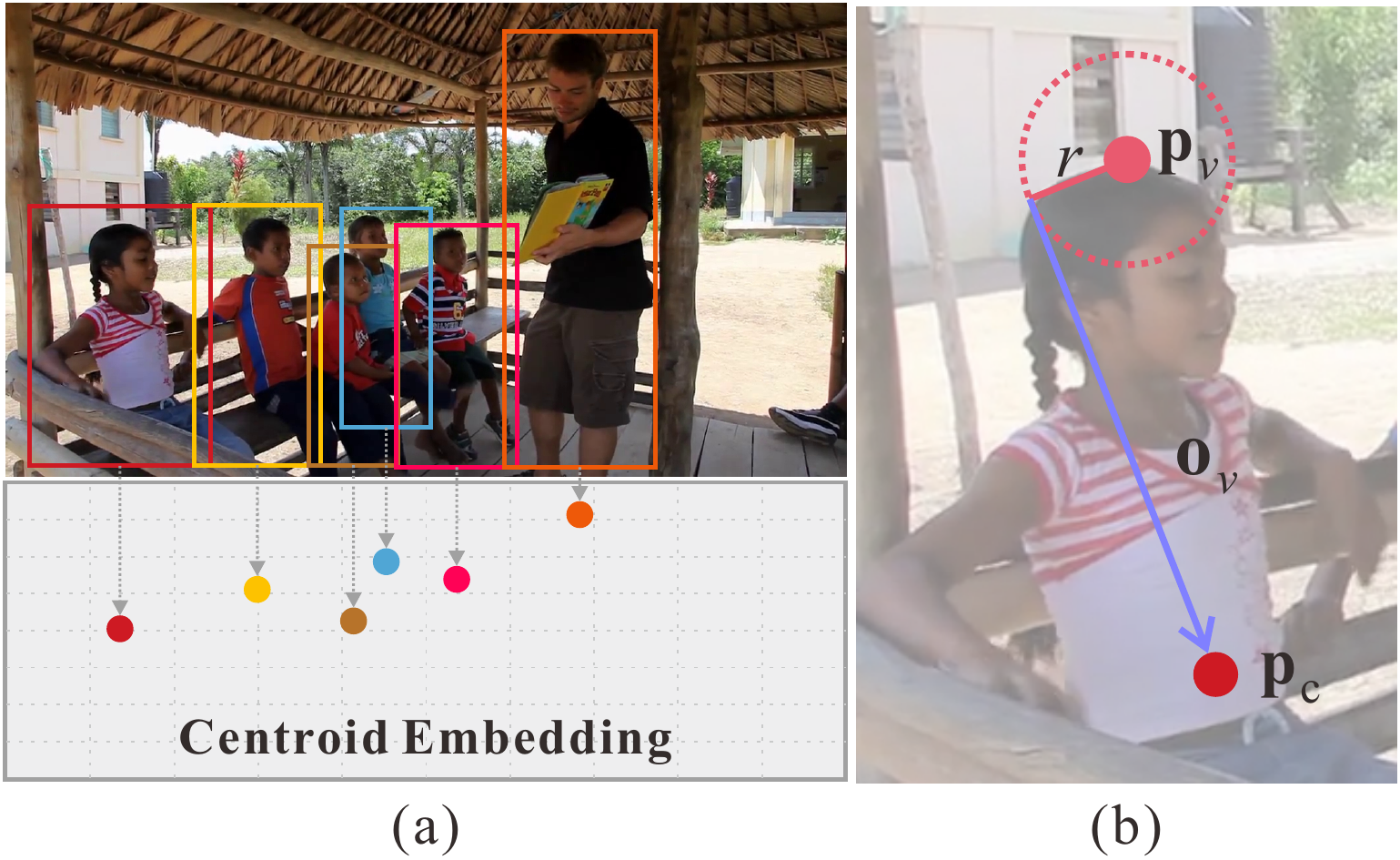}
\caption{Centroid embedding via dense joint regression. (a)  Centroid embedding results for persons in the image. (b) Construction of the regression target for a pixel in the image (Sec.~\ref{subsec:partition}).}
\label{fig:center_embedding_example}
\end{center}
\vspace{-20pt}
\end{figure}

\subsection{Joint Candidate Detection}
\label{subsec:joint_detection}

To reliably detect human body joints, we use  confidence maps to encode probabilities  of joints presenting at each position in the image. The joint confidence maps  are constructed by modeling
the joint locations as Gaussian peaks, as shown in Figure~\ref{fig:flowchart} (a). We use $\mathbf{C}_{j}$ to denote the confidence map for the $j$th joint with  $\mathbf{C}_{j}^i$ being the
confidence map of the $j$th joint for the $i$th person. For a position $\mathbf{p}_v$ in the given image, $\mathbf{C}_{j}^i(\mathbf{p}_v)$ is calculated by
$\mathbf{C}_{j}^i(\mathbf{p}_v){=}\exp\left(-{\|\mathbf{p}_v{-}\mathbf{p}_j^i\|_2^2}/{\sigma^2}\right)$,
where $\mathbf{p}_j^i$ denotes the groundtruth position of the $j$th joint of the $i$th person, and $\sigma$ is an empirically chosen constant to control  variance of the Gaussian distribution and set as 7 in the experiments.
The target confidence map, which the proposed GPN model learns to predict, is an aggregation of peaks of all the persons in a single map. Here, we choose to take the maximum of confidence maps
rather than average to remain distinctions between close-by peaks~\cite{cao2017realtime}, \ie $\mathbf{C}_{j}(\mathbf{p}_v){=}\max_i\mathbf{C}_{j}^i(\mathbf{p}_v)$.
During testing, we first find peaks with confidence scores greater than a given threshold $\tau$ (set as 0.1) on predicted confidence maps $\tilde{\mathbf{C}}$ for all types of joints.
Then we perform non-maximum suppression to find the joint candidate set $\tilde{\mathbf{p}}{=}\{\mathbf{p}_1,\mathbf{p}_2,\ldots, \mathbf{p}_N\}$.

\subsection{Joint Partition via Dense Regression}
\label{subsec:partition}

Our proposed joint partition model performs dense regression over all the joint candidates to localize centroids of multiple persons to partition joints into different person instances,
as shown in Figure~\ref{fig:flowchart} (b) and (c). It learns to transform all the pixels belonging to a specific person to an identical single point in a carefully deigned embedding space, where they are easy to cluster into corresponding persons. Such a dense regression framework enables generating joint partitions by one single feed-forward pass, reducing high joint detection complexity troubling top-down solutions.

To this end, we build and parameterize the joint candidate embedding space  by the person centroids, as centroid is  stable and reliable to discriminate  person instances even in presence of some extreme poses. We denote the constructed  embedding space as $\mathcal{H}$. In  $\mathcal{H}$, each person corresponds to a single point (\ie, the centroid), and  each point $\mathbf{h}_{*}{\in}\mathcal{H}$ represents a hypothesis about  centroid location of a specific person instance. An example is given in Figure~\ref{fig:center_embedding_example} (a).

Joint candidates are densely transformed into $ \mathcal{H} $ and can collectively determine  the centroid hypotheses of their corresponding person instances,  since they are tightly related in the view of articulated kinematics, as shown in Figure~\ref{fig:flowchart}~(c). For instance, a candidate of the head joint  would add votes for the presence of a person's centroid to the location just below it. A single candidate does not necessarily provide evidence for the exact centroid of a person instance, but the population of joint candidates can vote for the correct centroid with large probability and determine the joint partitions correctly.
 In particular, the probability of generating joint partition $\mathbf{g}_{*}$ at location $\mathbf{h}_{*}$ is calculated by summing the votes from different joint candidates together, \ie
\begin{equation*}
\begin{aligned}
&\mathbb{P}(\mathbf{g}_{*}|\mathbf{h}_{*}) \propto \\
&\hspace{2mm}\sum_j w_j\big(\sum_{\mathbf{p}_v \in \tilde{\mathbf{p}}}\mathbbm{1}[\tilde{\mathbf{C}}_{j}(\mathbf{p}_v) \geq \tau]\exp\{-\|f_j(\mathbf{p}_v)-\mathbf{h}_{*}\|_2^2\}\big),
\end{aligned}
\end{equation*}
where
$\mathbbm{1}[\cdot]$ is the indicator function and $w_j$ is the weight for the votes from $j$th joint category. We set $w_j{=}1$ for all joints assuming all kinds of joints equally contribute to the localization of person instances in view of unconstrained shapes of human body and uncertainties of presence of different joints. The  function $f_j{:}\mathbf{p} \rightarrow \mathcal{H}$ learns to densely transform  \emph{every} pixel in the image to  the embedding space $\mathcal{H}$. 
For learning $f_j$, we build  the target regression map $\mathbf{T}_{j}^i$  for the $j$th joint of the $i$th person as follows:
\begin{equation*}
\mathbf{T}_{j}^i(\mathbf{p}_v) = \left \{
\begin{array}{cl}
{\mathbf{o}_{j,v}^i}/{Z}   & \mbox{ if $\mathbf{p}_v \in \mathcal{N}_j^i$ }, \\
0                    & \mbox{ otherwise, }
\end{array} \right.
\end{equation*}
\begin{equation*}
\mathbf{o}_{j,v}^i=(\mathbf{p}_{\mathrm{c}}^i - \mathbf{p}_v)=(x_{\mathrm{c}}^i - x_v, y_{\mathrm{c}}^i - y_v),
\end{equation*}
where $\mathbf{p}_{\mathrm{c}}^i$ denotes the centroid position of the $i$th person, $Z{=}\sqrt{H^2+W^2}$ is the normalization factor, $H$ and $W$ are the height and width of image
$\mathbf{I}$, $\mathcal{N}_j^i{=}\{\mathbf{p}_v |~\|\mathbf{p}_v{-}\mathbf{p}_j^i\|_2{\leq}r \}$ denotes the neighbor positions of the $j$th joint of the $i$th person, and $r$ is
a constant to define the neighborhood size, set as 7 in our experiments. An example is shown in Figure~\ref{fig:center_embedding_example} (b) for construction of a regression target of a pixel in a given image.
Then, we define the target regression map $\mathbf{T}_{j}$ for the $j$th joint as the average for all persons by
\begin{equation*}
\mathbf{T}_{j}(\mathbf{p}_v) = \frac{1}{N_v}\sum_i \mathbf{T}_{j}^i(\mathbf{p}_v),
\end{equation*}
where $N_v$ is the number of non-zero vectors at position $\mathbf{p}_v$ across all persons. During testing, after predicting the  regression map $\tilde{\mathbf{T}}_{j}$,
we define transformation function $f_j$ for position $\mathbf{p}_v$ as $f_j(\mathbf{p}_v){=}\mathbf{p}_v + Z\tilde{\mathbf{T}}_{j}(\mathbf{p}_v)$.
After generating $\mathbb{P}(\mathbf{g}_{*}|\mathbf{h}_{*})$ for each point in the embedding space, we calculate the score $\varphi(\mathbf{p}, \mathbf{g})$ as $\varphi(\mathbf{p}, \mathbf{g}){=} \sum_{i}\log \mathbb{P}(\mathbf{g}_i|\mathbf{h}_i)$.

Then the problem of joint partition generation is converted to finding peaks in the  embedding space $\mathcal{H}$. As there are no  priors on the number of persons in the image,  we adopt the  Agglomerative Clustering~\cite{bourdev2009poselets} to find  peaks by clustering the  votes, which  can automatically determine the number of clusters. We denote the vote set as $\mathbf{h}{=}\{\mathbf{h}_v|\mathbf{h}_v{=}f_j(\mathbf{p}_v), \tilde{\mathbf{C}}_{j}(\mathbf{p}_v){\geq}\tau, \mathbf{p}_v{\in}\tilde{\mathbf{p}} \}$, and use $\mathcal{C}{=}\{\mathcal{C}_1,\ldots,\mathcal{C}_M\}$ to denote the clustering result on $\mathbf{h}$, where $\mathcal{C}_i$ represents the $i$th cluster and $M$ is the number of clusters. We assume the set of joint candidates casting votes in each cluster corresponds to a joint partition $\mathbf{g}_i$, defined by
\begin{equation}\label{eq:person_partition}
\mathbf{g}_i = \{\mathbf{p}_v | \mathbf{p}_v \in \tilde{\mathbf{p}}, \tilde{\mathbf{C}}_{j}(\mathbf{p}_v) \geq \tau, f_j(\mathbf{p}_v) \in \mathcal{C}_i\}.
\end{equation}

\begin{algorithm}[t!]\small
	\caption{Local greedy inference for multi-person pose estimation. }
	\label{alg:greedy_infer_alg}
	\SetKwInOut{Input}{input} \SetKwInOut{Output}{output}
		 \Input{joint candidates $\tilde{\mathbf{p}}$, person partitions $\tilde{\mathbf{g}}$, joint confidence maps $\tilde{\mathbf{C}}$, dense regression maps $\tilde{\mathbf{T}}$,  $\tau$.}
		 \Output{multi-person pose estimation $\mathcal{R}$}
		 \textbf{initialization:}  $\mathcal{R} \leftarrow \varnothing$ \\
		\For{ $\mathbf{g}_i \in \tilde{\mathbf{g}}$ }{
		\While{$\mathbf{g}_i \neq \varnothing$}{
		 Initialize single-person pose estimation  $\mathcal{P} \leftarrow \varnothing$\\
		\For{$j$th joint category, $j=1$ to $K$}{
		\eIf{$\mathcal{P} = \varnothing$}{
		 Find root joint candidate in $\mathbf{g}_i$ for $\mathcal{P}$ by: $\mathbf{p}_{*} \leftarrow \mathrm{arg}\max_{\mathbf{p}_v \in \mathbf{g}_i}\tilde{\mathbf{C}}_{j}(\mathbf{p}_v)$
		}
	{
	 Find joint candidate closest to  centroid $\mathbf{c}$: \\
	 $\mathbf{p}_{*} \leftarrow \mathrm{arg}\max_{\mathbf{p}_{v} \in \mathbf{g}_i} \frac{\mathbbm{1}[\tilde{\mathbf{C}}_{j}(\mathbf{p}_v) \geq \tau]}{\exp\{\|f_j(\mathbf{p}_v) - \mathbf{c}\|_2^2\}}$
	}
		\If{$\tilde{\mathbf{C}}_{j}(\mathbf{p}_{*}) \geq \tau$}{
         Update $\mathcal{P}{\leftarrow}\mathcal{P}{\cup}\{(\mathbf{p}_{*}, j)\}$, $\mathbf{g}_i{\leftarrow}\mathbf{g}_i{\setminus}\{\mathbf{p}_{*}\}$
		 Update $\mathbf{c}$ by averaging the person centroid hypotheses: $\mathbf{c} \leftarrow \sum_{(\mathbf{p}_v, n) \in \mathcal{P}}f_{n}(\mathbf{p}_v) / |\mathcal{P}|$\\
		
		}
			}
		 Update $\mathcal{R} \leftarrow \mathcal{R} \cup \{\mathcal{P}\}$
			}
		}
	\label{alg:local_greedy_search}
\end{algorithm}

\subsection{Local Greedy Inference for Pose Estimation}

According to Eqn.~(\ref{eq:gibbs_distribution}), we maximize the conditional probability $\mathbb{P}(\mathbf{p}, \mathbf{u}, \mathbf{b}, \mathbf{g} | \mathbf{I}, \Theta)$ by minimizing the energy function $E(\mathbf{p}, \mathbf{u}, \mathbf{b}, \mathbf{g})$ in Eqn.~(\ref{eq:energy_func}). We  optimize $E(\mathbf{p}, \mathbf{u}, \mathbf{b}, \mathbf{g})$ in two sequential steps: 1) generate joint partition set based on joint candidates;
2) conduct  joint configuration inference in each joint partition locally, which reduces the joint configuration complexity and overcomes the drawback of bottom-up approaches.

After getting  joint partition according to Eqn.~(\ref{eq:person_partition}), the score $\varphi(\mathbf{p}, \mathbf{g})$ becomes a constant. Let $\tilde{\mathbf{g}}$  denote the generated partition set. The optimization  is then simplified as 
\begin{equation}\label{eq:energy_func_sim}
\begin{aligned}
\tilde{\mathbf{u}},\tilde{\mathbf{b}}= \mathrm{arg}\min_{\mathbf{u},\mathbf{b}}\bigg(-\sum_{\mathbf{g}_i \in \tilde{\mathbf{g}}}
\Big( & \sum_{\mathbf{p}_v \in \mathbf{g}_i} \psi(\mathbf{p}_v, u_v) + \\
&\hspace{-12mm}\sum_{\mathbf{p}_v,\mathbf{p}_w \in \mathbf{g}_i}\phi(\mathbf{p}_v, u_v, \mathbf{p}_w, u_w)\Big)\bigg).
\end{aligned}
\end{equation}
Pose estimation in each joint partition is independent, thus inference over different joint partitions becomes separate. We propose the following local greedy inference algorithm to solve
Eqn.~(\ref{eq:energy_func_sim})
for multi-person pose estimation. Given a joint partition $\mathbf{g}_i$,   the unary term $\psi(\mathbf{p}_v, u_v)$ is the confidence score   at $\mathbf{p}_v$  from the $u_v$th joint detector: $\psi(\mathbf{p}_v, u_v){=}\tilde{\mathbf{C}}_{u_v}(\mathbf{p}_v)$.
The binary term $\phi(\mathbf{p}_v, u_v, \mathbf{p}_w, u_w)$ is  the similarity score of votes of two joint candidates based on the global affinity cues in the  embedding space:
\begin{equation*}
\begin{aligned}
&\phi(\mathbf{p}_v, u_v, \mathbf{p}_w, u_w) \\
&= \mathbbm{1}[\tilde{\mathbf{C}}_{u_v}(\mathbf{p}_v) \geq \tau]\mathbbm{1}[\tilde{\mathbf{C}}_{u_w}(\mathbf{p}_w) \geq \tau]\exp\{-\|\mathbf{h}_{v}-\mathbf{h}_{w}\|_2^2\},
\end{aligned}
\end{equation*}
where $\mathbf{h}_v{=}\mathbf{p}_v{+}Z\tilde{\mathbf{T}}_{u_v}(\mathbf{p}_v)$ and $\mathbf{h}_w{=}\mathbf{p}_w{+}Z\tilde{\mathbf{T}}_{u_w}(\mathbf{p}_w)$.

For efficient inference in Eqn.~(\ref{eq:energy_func_sim}), we adopt a greedy strategy which guarantees the energy monotonically decreases and eventually converges to a lower bound.
Specifically, we iterate through each joint one by one, first considering joints around  torso and  moving out to  limb. We start the inference with  neck.
For a  neck candidate, we use its embedding point in  $ \mathcal{H} $ to initialize the  centroid of its person instance.
Then, we select the head top candidate closest to the person centroid and associate it with the same person as the neck candidate.
After that, we update  person centroid  by averaging the derived hypotheses. We loop through all other joint candidates  similarly.
Finally,  we get a person instance and its associated joints.  After utilizing neck as root for inferring joint configurations of person instances, if some candidates remain unassigned,
we utilize joints from torso, then from limbs, as the root to infer the person instance.  After all candidates find their associations to persons, the inference terminates. See details in Algorithm~\ref{alg:local_greedy_search}.

\begin{figure}[t!]
\begin{center}
\includegraphics[scale=0.47]{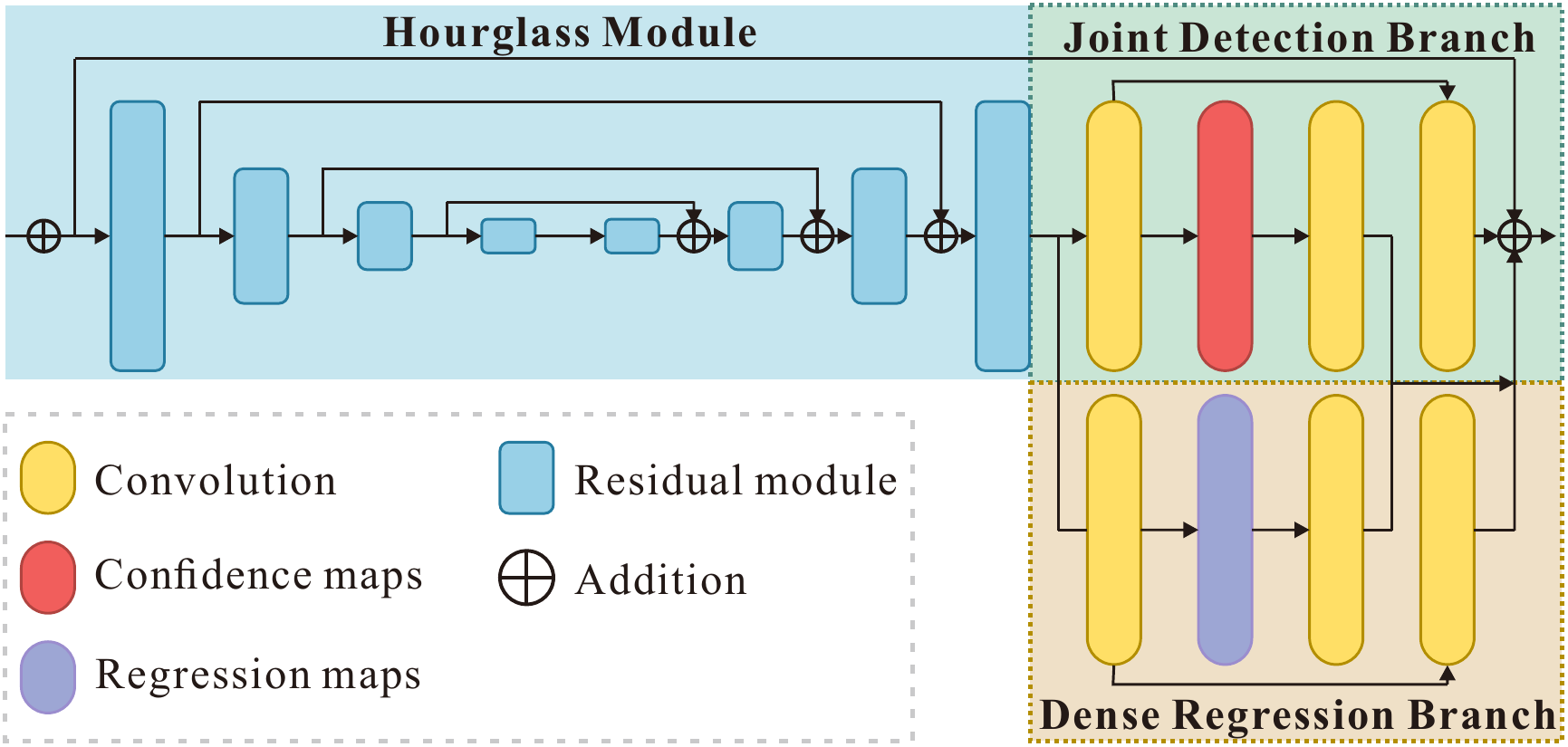}
\caption{Architecture of Generative Partition Network.   Its backbone is an Hourglass module (in blue block), followed by two branches:  joint detection (in green block) and  dense regression for joint partition (in yellow block). 
}
\label{fig:gpn_network_arch}
\end{center}
\vspace{-20pt}
\end{figure}

\section{Learning Joint Detector and Dense Regressor with CNNs}


GPN is a generic model and compatible with various CNN architectures. Extensive  architecture engineering is out of the scope of this work.  We simply choose the state-of-the-art Hourglass network~\cite{hpe:hourglass_arxiv15} as the backbone of GPN. Hourglass network consists of a sequence of Hourglass modules. As shown in Figure~\ref{fig:gpn_network_arch}, each Hourglass module first learns down-sized   feature maps from the input image, and then recovers full-resolution feature maps through up-sampling for precise joint localization. In particular, each Hourglass module is implemented as a fully convolutional network. Skipping connections are added between feature maps with the same resolution symmetrically to capture information at every scale. Multiple Hourglass modules are stacked sequentially for gradually refining the predictions via reintegrating the previous estimation results. Intermediate supervision is applied at each Hourglass module.

Hourglass network was  proposed for \emph{single}-person pose estimation. GPN extends it to \emph{multi}-person cases.  GPN introduces modules enabling simultaneous joint detection (Sec.~\ref{subsec:joint_detection})  and dense joint-centroid regression (Sec.~\ref{subsec:partition}), as shown in Figure~\ref{fig:gpn_network_arch}. In particular, GPN utilizes the Hourglass module to learn image  representations and then separates into two branches: one  produces the dense regression maps for detecting person centroids,   via one $3{\times}3$ convolution on feature maps from the Hourglass module and another  $1{\times}1$ convolution for classification; the other branch produces  joint detection confidence maps. With this design, GPN obtains joint detection and partition in one feed-forward pass. When using multi-stage Hourglass modules, GPN feeds the predicted dense regression maps  at every stage into the next one through $1{\times}1$ convolution, and then combines  intermediate features with  features from the previous stage.

For training GPN, we use $\ell_2$ loss to learn both joint detection and dense regression branches with supervision at each stage. The losses are defined as
\begin{equation*}
\begin{aligned}
&L_{\text{joint}}^t  \triangleq \sum_{j}\sum_{v}\|\tilde{\mathbf{C}}_{j}^t(\mathbf{p}_v)-\mathbf{C}_{j}(\mathbf{p}_v)\|_2^2 \ \ \ \\
&L_{\text{regression}}^t       \triangleq \sum_{j}\sum_{v}\|\tilde{\mathbf{T}}_{j}^t(\mathbf{p}_v)-\mathbf{T}_{j}(\mathbf{p}_v)\|_2^2,
\end{aligned}
\vspace{-3mm}
\end{equation*}
where $\tilde{\mathbf{C}}_{j}^t$ and $\tilde{\mathbf{T}}_{j}^t$ represent predicted joint confidence maps and dense regression maps at the $t$th stage, respectively.
The groundtruth $\mathbf{C}_{j}(\mathbf{p}_v)$ and $\mathbf{T}_{j}(\mathbf{p}_v)$ are constructed as  in  Sec.~\ref{subsec:joint_detection} and~\ref{subsec:partition} respectively.
The total loss is given by
$L{=}\sum_{t=1}^T(L_{\text{joint}}^t + \alpha L_{\text{regression}}^t)$,
where $T{=}8$ is the number of  Hourglass modules (stages) used in our implementation  and  the weighting factor $\alpha$  is empirically set as $1$.

\section{Experiments}

\subsection{Experimental Setup}

\paragraph{Datasets} We evaluate the proposed GPN  on three widely adopted benchmarks: MPII Human Pose Multi-Person (MPII) dataset~\cite{andriluka14cvpr}, extended PASCAL-Person-Part dataset~\cite{xia2017joint}, and ``We Are Family'' (WAF) dataset~\cite{eichner2010we}. The MPII dataset consists of 3,844  and 1,758  groups of multiple interacting persons for training and testing respectively. Each person in the image is annotated for 16 body joints. It also provides  more than 28,000 training samples for single-person pose estimation. The extended PASCAL-Person-Part dataset contains 3,533 challenging images from the original PASCAL-Person-Part dataset~\cite{chen2014detect}, which are split into 1,716  for training and 1,817  for testing. Each person is annotated with 14 body joints shared with MPII dataset, without pelvis and thorax.  The WAF dataset contains 525 web images (350   for training and 175  for testing). Each person  is annotated with 6 line segments for the upper-body.

\vspace{-4mm}
\paragraph{Data Augmentation} We follow  conventional ways to augment training samples by cropping original images based on the  person center. In particular, we augment each training sample with rotation degrees sampled in $[-40^{\circ}, 40^{\circ}]$, scaling  factors in $[0.7, 1.3]$, translational offset in $[-40\text{px}, 40\text{px}]$ and horizontally mirror. We resize each training sample to  $256 {\times} 256$ pixels with padding.

\vspace{-4mm}
\paragraph{Implementation} For  MPII dataset, we reserve 350 images randomly selected from the training set for validation. We use the rest training images and all the provided single-person  samples to train the GPN for 250 epochs. For evaluation on the other two datasets, we  follow the common practice and finetune the GPN model pretrained on MPII for 30 epochs. To deal with some extreme cases where centroids of persons are overlapped, we slightly perturb the centroids by adding small offset to separate them. We implement our model with PyTorch~\cite{paszke2017pytorch} and adopt the RMSProp~\cite{rmsprop2012} for optimization. The initial learning rate is 0.0025 and decreased by multiplying 0.5 at the 150th, 170th, 200th, 230th epoch. In testing, we follow conventions to crop image patches using the given position and  average person scale of test images, and resize and pad the cropped samples to $384{\times}384$ as input to GPN. We search for suitable image scales  over 5 different choices. Specially, when testing on  MPII, following previous works~\cite{cao2017realtime,newell2016associative}, we apply a single-person model~\cite{hpe:hourglass_arxiv15} trained on  MPII  to  refine the estimations. We use the standard Average Precision (AP) as performance metric on all the  datasets, as suggested by~\cite{hpe:deepercut_eccv16,xia2017joint}.  We will make
codes and pre-trained models available.

\begin{table}[t!]\scriptsize
  \caption{Comparison with state-of-the-arts on the full testing set of MPII Human Pose Multi-Person dataset (AP).}
  \label{tab:exp_mpii_sota}
  \centering
  \setlength{\tabcolsep}{2pt}
  \begin{tabular}{lccccccccc}
    \toprule
    Method   &Head & Sho. & Elb. & Wri. & Hip & Knee  & Ank. & Total & Time [s]\\
    \midrule
    Iqbal and Gall~\cite{iqbal2016multi} & 58.4  & 53.9  & 44.5  & 35.0  & 42.2  & 36.7 & 31.1 & 43.1 & 10\\
    Insafutdinov \emph{et al.}~\cite{hpe:deepercut_eccv16} & 78.4  & 72.5  & 60.2  & 51.0  & 57.2  & 52.0 & 45.4 & 59.5 & 485\\
    Levinkov \emph{et al.}~\cite{levinkov2017joint} & 89.8 & 85.2 &	71.8 &	59.6 &	71.1 &	63.0 &	53.5 &	70.6 & - \\
    Insafutdinov \emph{et al.}~\cite{insafutdinov2016articulated} & 88.8  & 87.0  & 75.9  & 64.9  & 74.2  & 68.8 & 60.5 & 74.3 & - \\
    Cao \emph{et al.}~\cite{cao2017realtime} & 91.2  & 87.6  & 77.7  & 66.8  & 75.4  & 68.9 & 61.7 & 75.6 & 1.24\\
    Fang \emph{et al.}~\cite{fang16rmpe}& 88.4  & 86.5  & 78.6  & 70.4  & 74.4  & 73.0 & 65.8 & 76.7 & 1.5\\
    Newell and Deng~\cite{newell2016associative} & 92.1 & 89.3 & 78.9 & 69.8 & 76.2 & 71.6 & 64.7 & 77.5 & - \\
    \midrule
    GPN (Ours) & \textbf{92.2}  & \textbf{89.7}  & \textbf{82.1}  & \textbf{74.4}  & \textbf{78.6}  & \textbf{76.4} & \textbf{69.3} & \textbf{80.4} & \textbf{0.77} \\
    \bottomrule
  \end{tabular}
\end{table}

\begin{table}[t!]\footnotesize
    \caption{Comparison with state-of-the-arts on the testing set of the extended PASCAL-Person-Part dataset (AP)}
    \label{tab:exp_pascal_sota}
    \centering
    \setlength{\tabcolsep}{2pt}
    \begin{tabular}{lcccccccccc}
        \toprule
        Method &Head & Sho. & Elb. & Wri. & Hip & Knee  & Ank. & Total\\
        \midrule
        Chen and Yuille~\cite{chen2015parsing} & 45.3 & 34.6 & 24.8 & 21.7 & 9.8 & 8.6 & 7.7 &  21.8\\
        Insafutdinov et al.~\cite{hpe:deepercut_eccv16} & 41.5 & 39.3 & 34.0 & 27.5 & 16.3 & 21.3 & 20.6 &  28.6\\
        Xia et at.~\cite{xia2017joint} & 58.0 & 52.1 & 43.1 & 37.2 & 22.1 & 30.8 & 31.1 & 39.2\\
        \midrule
        GPN (Ours) & \textbf{66.9} & \textbf{60.0} & \textbf{51.4} & \textbf{48.9} & \textbf{29.2} & \textbf{36.4} & \textbf{33.5}  & \textbf{46.6} \\
        \bottomrule
    \end{tabular}
\end{table}

\begin{table}[t!]\footnotesize
  \caption{Comparison with state-of-the-arts on the testing set of the WAF dataset (AP).}
  \label{tab:exp_waf_sota}
  \centering
  \setlength{\tabcolsep}{5pt}
  \begin{tabular}{lccccc}
    \toprule
    Method   &Head & Shoulder & Elbow & Wrist & Total\\
    \midrule
    Chen and Yuile~\cite{chen2015parsing} & 83.3 & 56.1 & 46.3 & 35.5 & 55.3 \\
    Pishchulin \emph{et al.}~\cite{hpe:deepcut_cvpr16} & 76.6 & 80.8 & 73.7 & 73.6 & 76.2 \\
    Insafutdinov \emph{et al.}~\cite{hpe:deepercut_eccv16} & 92.6 & 81.1 & 75.7 & 78.8 & 82.0 \\
    \midrule
    GPN (Ours) & \textbf{93.1} & \textbf{82.9} & \textbf{83.5} & \textbf{79.9} & \textbf{84.8}\\
    \bottomrule
  \end{tabular}
\end{table}

\subsection{Results and Analysis}

\paragraph{MPII} Table~\ref{tab:exp_mpii_sota} shows the evaluation results on the full testing set of MPII. We can see that the proposed GPN achieves overall $80.4\%$ AP and significantly outperforms previous state-of-the-art achieving $77.5\%$ AP~\cite{newell2016associative}. In addition, the proposed GPN improves the performance for localizing all the joints consistently. In particular, it brings remarkable improvement over rather difficult joints mainly caused by occlusion and high degrees of freedom, including wrists ($74.4\%$ vs $69.8\%$ AP), ankles ($69.3\%$ vs $64.7\%$ AP), and knees (with absolute $4.8\%$ AP increase over~\cite{newell2016associative}), confirming the robustness of the proposed generative model and global affinity cues to these distracting factors. These results clearly show  GPN is outstandingly effective for multi-person pose estimation. We also report the computational speed of GPN\footnote{The runtime time is measured on CPU Intel I7-5820K 3.3GHz and GPU TITAN X (Pascal). The time is counted with 5 scale testing, not including the refinement time by single-person pose estimation.}   in Table~\ref{tab:exp_mpii_sota}.  GPN is about 2 times faster than the bottom-up approach~\cite{cao2017realtime} with state-of-the-art speed for multi-person pose estimation.
This demonstrates the efficiency of performing joint detection and partition simultaneously in our model.

\vspace{-4mm}
\paragraph{PASCAL-Person-Part} Table~\ref{tab:exp_pascal_sota} shows the evaluation results. GPN provides absolute $7.4\%$ AP improvement ($46.6\%$ vs $39.2\%$ AP) over the state-of-the-art~\cite{xia2017joint}. Moreover, the proposed GPN brings significant improvement on difficult joints, such as wrist  ($48.9\%$ vs $37.2\%$ AP). These results further demonstrate the effectiveness and robustness of our model for multi-person pose estimation.

\vspace{-4mm}
\paragraph{WAF} As shown in Table~\ref{tab:exp_waf_sota}, GPN  achieves overall $84.8\%$ AP, bringing  $3.4\%$ improvement over the best bottom-up approach~\cite{hpe:deepercut_eccv16}. GPN achieves the best performance for all upper-body joints. In particular, it gives the most significant performance  improvement on the elbow, about $10.3\%$ higher than previous best results. These results  verify the effectiveness of the proposed GPN for tackling the multi-person pose estimation problem.

\begin{figure*}[t!]
\begin{center}
\includegraphics[scale=0.725]{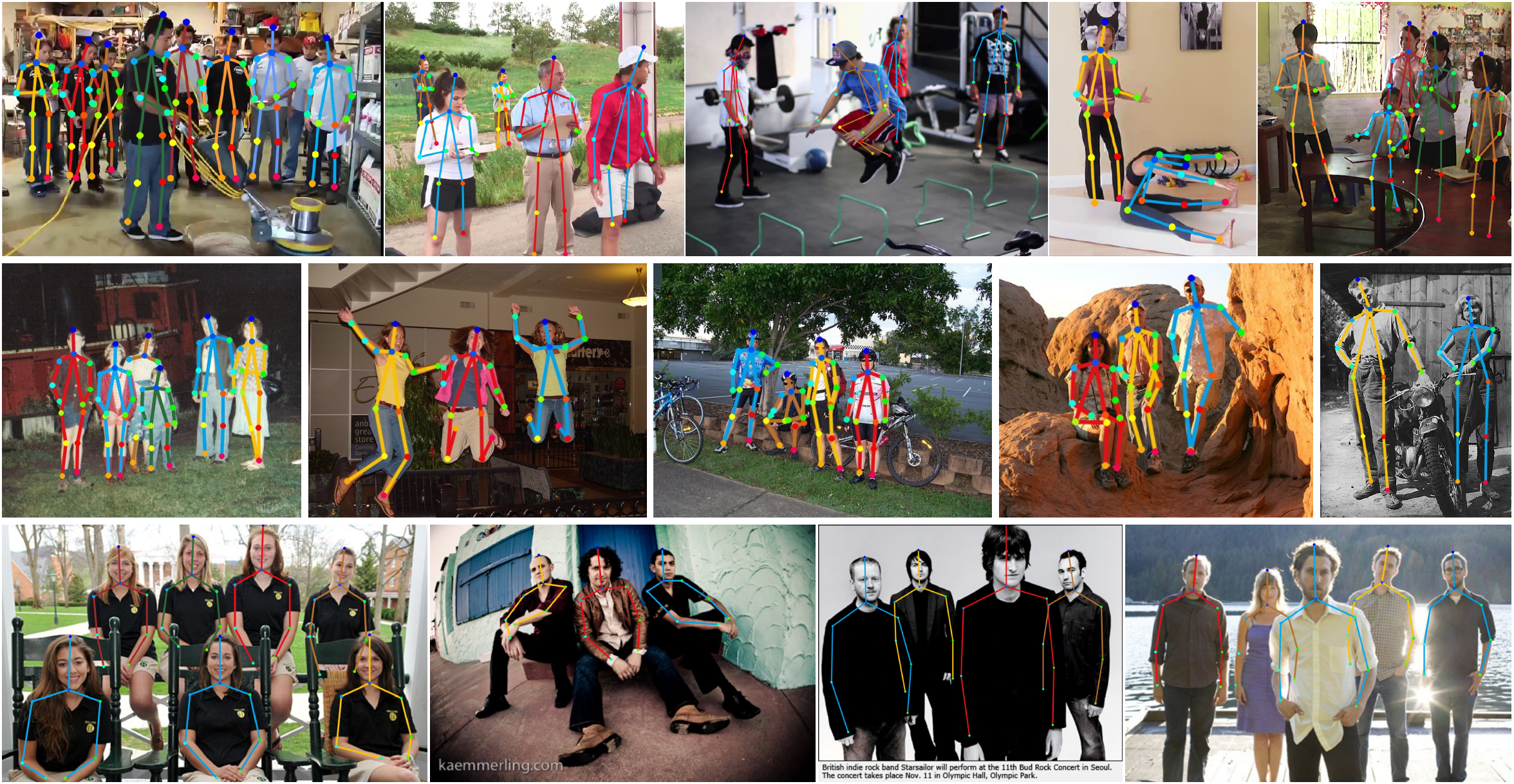}
\caption{Qualitative results of the proposed GPN on  MPII (1st row), extended PASCAL-Person-Part  (2nd row), and  WAF (3rd row). GPN performs well  even in  challenging scenarios, \emph{e.g.}, self-occlusion and other overlapped persons  in the 1st example of 1st row,
large pose variations  in the 2nd example of 2nd row, appearance and illumination changes in the 4th example of  3rd row.}
\label{fig:mpi_vis}
\end{center}
\vspace{-20pt}
\end{figure*}

\vspace{-4mm}
\paragraph{Qualitative Results} Figure~\ref{fig:mpi_vis} visualizes some pose estimation results on these three datasets. We can observe that the proposed GPN model estimates multi-person poses accurately and robustly even in challenging scenarios, \eg, joint occlusion caused by a person of interest and other overlapped persons presenting in the first example of MPII dataset, large pose variation shown in the second example of the extended PASCAL-Person-Part dataset, and appearance and illumination changes in the forth example of WAF dataset. These results also verify the effectiveness of GPN for producing reliable joint detections and partitions in multi-person pose estimation. 

\begin{table}[t!]\scriptsize
  \caption{Ablation experiments on MPII validation set (AP).}
  \label{tab:ablation_analysis}
  \centering
  \setlength{\tabcolsep}{1.5pt}
  \begin{tabular}{lccccccccc}
    \toprule
     Method   &Head & Sho. & Elb. & Wri. & Hip & Knee  & Ank. & Total & InferTime [ms]\\
     \midrule
     GPN-Full & 94.4 & 90.0     & 81.3  & 72.1 & 77.8 & 72.7 & 64.7  & 79.0 & 1.9\\
     \midrule
     GPN-w/o-Partition & 93.2 & 89.3     & 79.9  & 70.1 & 78.8 & 73.1 & 65.7  & 78.6  & 3.4\\
     GPN-w/o-LGI & 93.1 & 89.1     & 79.5  & 68.5 & 79.0 & 71.4 & 64.4  & 77.8 & - \\
     GPN-w/o-Refinement & 90.4 & 86.8     & 79.3  & 69.8 & 77.5 & 69.3 & 61.9  & 76.4 & -\\
     GPN-$256{\times}256$ & 91.0 & 87.1     & 78.6  & 70.2 & 76.7 & 70.5 & 60.0  & 76.3 & -\\
     GPN-Vanilla & 90.5 & 86.4 & 77.1  & 69.4 & 72.2 & 67.7 & 60.2  & 74.8 & - \\
     \bottomrule
  \end{tabular}
\end{table}


\subsection{Ablation Analysis}

We conduct ablation analysis for the proposed GPN model using the  MPII validation set.  We evaluate multiple variants of our proposed GPN model by removing certain components from the full model (``GPN-Full'').  ``GPN-w/o-Partition''   performs inference on the whole image without using obtained joint partition information, which is similar to the pure bottom-up approaches.  ``GPN-w/o-LGI'' removes the local greedy inference phase. It allocates joint candidates to persons through finding the most activated  position for each  joint in each joint partition. This is similar to the top-down approaches.   ``GPN-w/o-Refinement'' does not perform refinement by using single-person pose estimator. We use ``GPN-$256{\times}256$'' to denote  testing over $256{\times}256$  images and ``GPN-Vanilla''  to denote single scale testing without refinement.

From  Table \ref{tab:ablation_analysis}, ``GPN-Full'' achieves  $79.0\%$ AP and the joint partition inference  only costs 1.9ms, which is very efficient. ``GPN-w/o-Partition'' achieves slightly lower AP ($78.6\%$)  with slower inference speed (3.4ms). The results confirm  effectiveness of generating joint partitions by  GPN\textemdash  inference within each joint partition individually reduces complexity and improves pose estimation  over multi-persons. Removing the local greedy inference phase as in ``GPN-w/o-LGI'' decreases the performance  to $77.8\%$ AP, showing local greedy inference is beneficial to pose estimation by effectively handling  false alarms of joint candidate detection based on global affinity cues in the embedding space. Comparison of ``GPN-w/o-Refinement''($76.4\%$ AP) with the full model  demonstrates that single-person pose estimation can refine  joint localization. ``GPN-Vanilla'' achieves $74.8\%$ AP, demonstrating the stableness of the proposed approach for multi-person pose estimation even in the case of removing  refinement and multi-scale testing.

\begin{figure}[t!]
\begin{center}
\setlength{\tabcolsep}{3.5pt}
\begin{tabular}{ccc}
\includegraphics[scale=0.21]{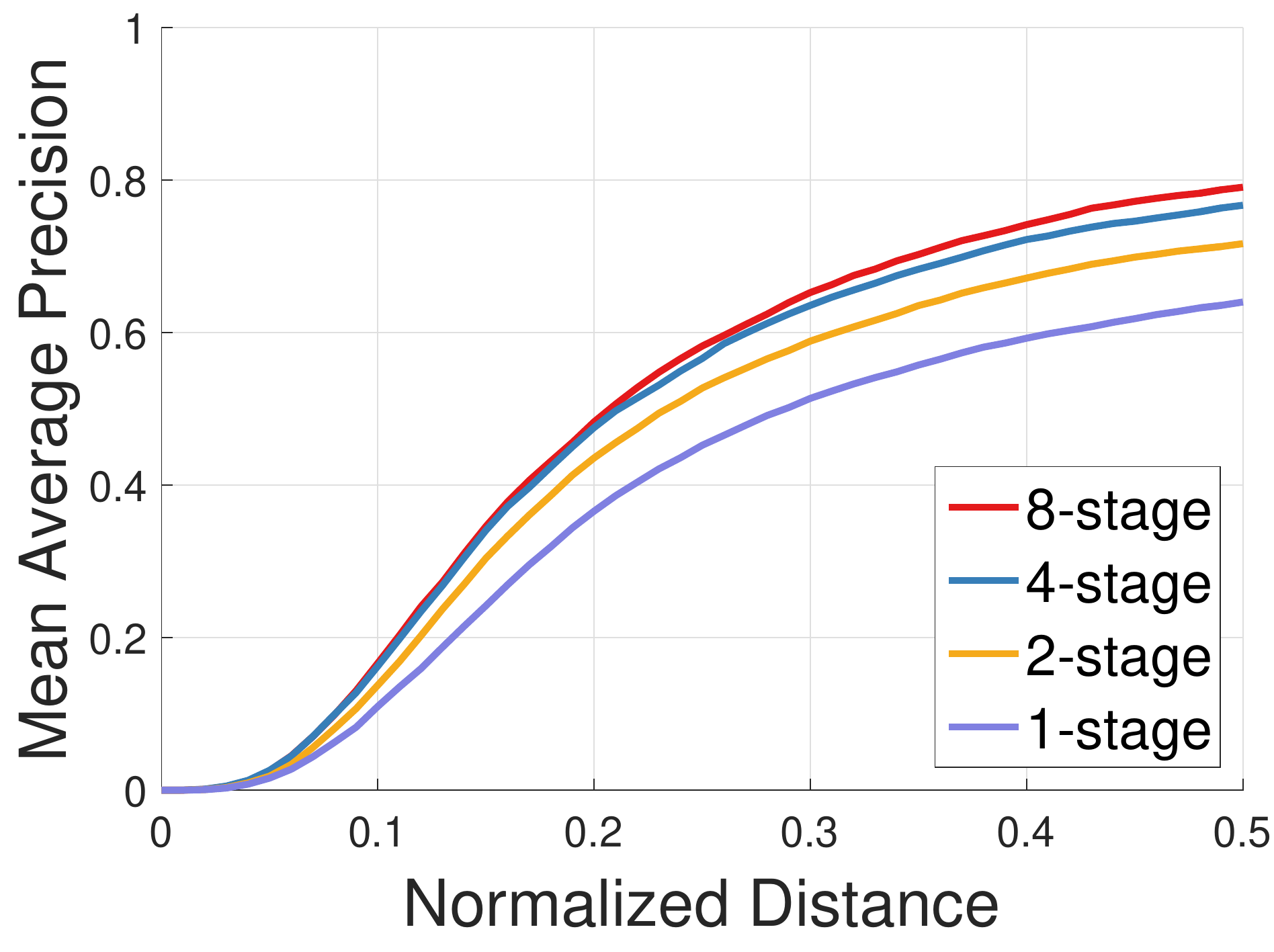} & \includegraphics[scale=0.26]{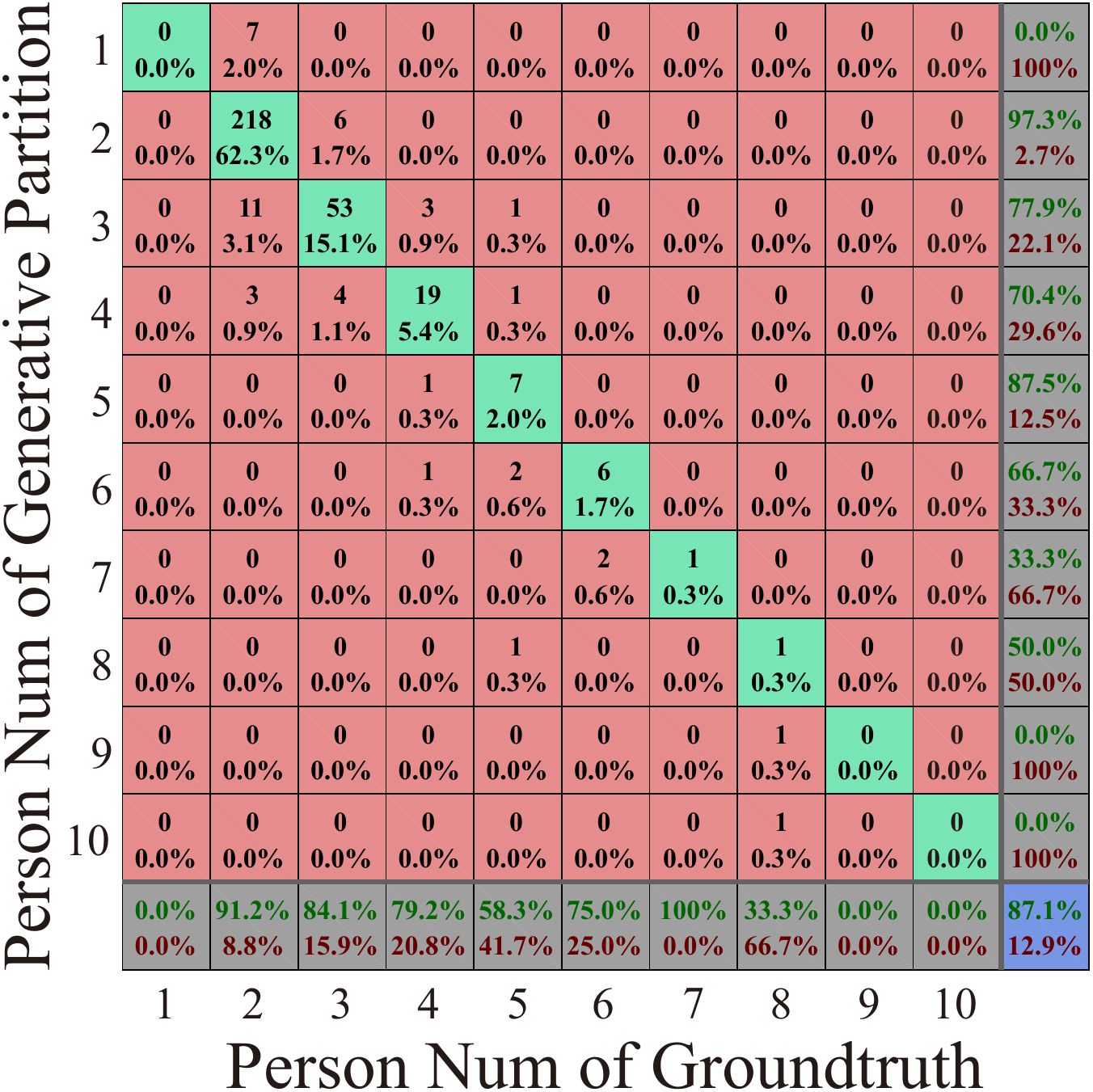} \\
{\fontsize{9pt}{9pt} \selectfont (a) } & {\fontsize{9pt}{9pt} \selectfont (b) } \\
\end{tabular}
\caption{(a) Ablation study on multi-stage Hourglass network. (b) Confusion matrix on person number inferred from generative partition (Sec.~\ref{subsec:partition}) with groundtruth. Mean square error is $0.203$. Best viewed in color and $2\times$ zoom.}
\label{fig:other_ablation_exp}
\vspace{-15pt}
\end{center}
\end{figure}

We also evaluate the pose estimation results from 4 different stages of the GPN model and plot the results in Figure~\ref{fig:other_ablation_exp} (a).   The performance increases monotonically when traversing more stages. The final results achieved at the $8$th stage give about $23.4\%$ improvement comparing with the first stage ($79.0\%$ vs $64.0\%$ AP). This is because the proposed GPN can recurrently correct errors on the dense regression maps along with the joint confidence maps conditioned on previous estimations in the multi-stage design, yielding gradual improvement on the joint
detections and partitions for multi-person pose estimation.

Finally, we evaluate the effectiveness of generative partition model for partition  person instances. In particular, we evaluate how its produced partitions match the real number of persons. The confusion matrix is shown in Figure~\ref{fig:other_ablation_exp} (b). We can observe that the proposed generative partition model can predict very close number of persons with the groundtruth, with   mean square error as small as $0.203$.

\section{Conclusion}

We  presented the Generative Partition Network (GPN) to efficiently and effectively address the challenging multi-person pose estimation problem. GPN solves the problem by simultaneously detecting and partitioning joints for multiple persons. It introduces a new approach to generate partitions through inferring over joint candidates in the  embedding space parameterized by person centroids. Moreover, GPN introduces a local greedy inference approach to estimate poses for person instances by utilizing the partition information. We demonstrate that GPN can provide appealing efficiency for both joint detection and  partition, and it can significantly overcome limitations of pure top-down and bottom-up solutions on three benchmarks multi-person pose estimation datasets.

{\small
\bibliographystyle{ieee}
\bibliography{ieee}
}

\end{document}